\documentclass{article}
\usepackage[T1]{fontenc}
\usepackage{amssymb,amsmath}
\usepackage{txfonts}
\usepackage{microtype}
\usepackage{xspace}
\xspaceaddexceptions{\%}
\usepackage{ctable}

\usepackage{paralist}

\usepackage{graphicx}
\usepackage{subfig} 

\usepackage{natbib}

\usepackage{algorithm}
\usepackage{algorithmic}

\usepackage[hyphens]{url}
\usepackage{hyperref}

\usepackage{icml-adapted}
\def\projectTitle{BaIT: Barometer for Information Trustworthiness}
\def\Names{Ois\'in Nolan, Jeroen van Mourik, Callum Rhys Tilbury}

\newcommand\blfootnote[1]{%
  \begingroup
  \renewcommand\thefootnote{}\footnote{#1}%
  \addtocounter{footnote}{-1}%
  \endgroup
}

\begin{document} 

\twocolumn[
\mlptitle{\projectTitle}

\centerline{\Names}

\centerline{\emph{School of Informatics, University of Edinburgh}}

\vskip 5mm
]
\blfootnote{All authors contributed equally.}

\begin{abstract}
This paper presents a new approach to the FNC-1 fake news classification task which involves employing pre-trained encoder models from similar NLP tasks, namely sentence similarity and natural language inference, and two neural network architectures using this approach are proposed. Methods in data augmentation are explored as a means of tackling class imbalance in the dataset, employing common pre-existing methods and proposing a method for sample generation in the under-represented class using a novel sentence negation algorithm. Comparable overall performance with existing baselines is achieved, while significantly increasing accuracy on an under-represented but nonetheless important class for FNC-1.
\end{abstract} 

\section{Introduction}
\label{sec:introduction}
In the modern age, where most people rely on online platforms for their information, the rapid spreading of fake news poses a significant problem \cite{vosoughi_spread_2018}. Consequently, the challenge of automatic fake news detection has attracted the attention of machine learning and linguistics researchers, and several approaches have been proposed (see e.g.\ \citealp{busioc_literature_2020} for an overview of NLP-based approaches). In reality, however, the `truth labelling' of claims is incredibly complex, even for humans, and such fully automated detection systems can thus hardly be implemented in practice. A more feasible alternative would be reliable AI systems assisting humans in recognizing harmful spreading of fake news. Specifically, a useful first step is to classify the relationship of the headline and body of the news article -- a task known as stance detection. This would enable a human reader to quickly extract a number of articles that agree, discuss, or disagree with a certain text and use this to make their own judgement.

From this perspective, the Fake News Challenge (FNC-1) was organized in 2017 to encourage the development of artificial intelligence tools to combat fake news and assist human fact checkers\footnote{\url{http://www.fakenewschallenge.org/}}. To this end, they provide a data set containing headlines and bodies of news articles, each pair of which is labelled as \textit{agree}, \textit{disagree}, \textit{discuss}, or \textit{unrelated}. The aim is to classify the stance of a given headline w.r.t.\ a given body.

In this paper, we develop a novel approach to the \mbox{FNC-1} stance classification task that utilizes pre-trained transformer models, which are extended with a custom attention layer and a neural classifier. Large transformer-based language models are a recent development in NLP which have proven to be massively successful in a wide variety of tasks \cite{devlin_bert_2018}. In our approach, we combine transformer models that have been trained on two domains: Natural Language Inference (NLI) and Sentence Similarity (SIM). These models are applied to encode sentences into vector representations that contain information useful for solving the corresponding task. We perform the classification in a hierarchical manner, first predicting whether a head is \textit{unrelated} to the body, and, if not, classifying its stance as one of the remaining three (related) categories: \textit{agree}, \textit{disagree}, or \textit{discuss}. The second task is more challenging than the first, as it requires identifying logical relations between head-body pairs, which are often subtle in the text. Pre-trained models were chosen based on the apparent similarity in the nature of the pre-train objective and the tasks at hand: SIM embeddings are used to discriminate between related and unrelated pairs, and NLI embeddings to classify the logical relationship. NLI was chosen here as its task formulation is remarkably similar to that of the second sub-task in FNC-1, namely classifying the relationship between a pair of sentences as one of \textit{entailment}, \textit{contradiction}, or \textit{neutrality}. In the most extensive model, an attention layer is applied that weighs NLI embeddings of the body sentences based on their similarity w.r.t.\ to head, computed from SIM embeddings. It is noteworthy that while the \textit{disagree} class is particularly important in this task (it is the case in which the news is `fake'), it is severely under-represented in the dataset, accounting for only 2\% of the samples. As a result of this imbalance, many existing works have scored impressive overall performance on the task with very poor accuracy on the \textit{disagree} class. In light of this observation, this work places particular emphasis on the \textit{disagree} accuracy, and tackling imbalance in the dataset.

Finally, the main contributions of this paper can be summarized as follows:
\begin{enumerate}[(i)]
    \item The proposal of a novel approach for stance detection in the news domain leveraging the power of large transformer models trained on NLI and SIM tasks.
    \item The development of a neural-network architecture that intuitively combines these two representation types, using an attention layer, to effectively transfer knowledge from the NLI and SIM domains to the problem of stance detection.
    \item An evaluation of methods in data augmentation as a means of resolving issues of class imbalance in the FNC-1 dataset. Methods discussed include common pre-existing methods as well as a novel method for creating synthetic data samples.
\end{enumerate}

The remainder of this paper is structured as follows. Section~\ref{sec:related work} provides a concise review of the related literature. Section~\ref{sec:data} discusses in more detail the relevant task and the data set used. Section~\ref{sec:methodology} extensively presents our proposed approach to the problem. Section~\ref{sec:experiments} describes the experiments designed to analyze the effectiveness of our method and discusses the results. Finally, Section~\ref{sec:conclusion} summarizes the most important findings and concludes with some recommendations for further research.

\section{Literature Review
}
\label{sec:related work}
As described above, this paper studies a variant of the stance detection problem.
Earlier work on this task frequently focused on target-specific stance classification, which is concerned with determining the attitude of a task towards a specified target, e.g.\ a person. Such work considered various domains, including debates \cite{hasan_stance_2013}, student essays \cite{faulkner_automated_2014}, or tweets \cite{augenstein_stance_2016}. The latter proposes an method based on text encoding using an LSTM architecture \cite{hochreiter1997long}, which is a popular approach in stance detection problems. In 2016, the Emergent dataset was introduced \cite{ferreira_emergent_2016}, which comprises news articles relating to 300 rumoured claim. The task is to determine whether the headline of an article was for, against or observing the associated claim, which is closer to our problem. The authors propose a simple logistic regression classifier based on hand-crafted features extracted from both the headline and the claim. For a more comprehensive overview on the available literature on stance detection, the reader is referred to the recent survey by \citealp{kucuk_stance_2020}.

The FNC-1 data set is a modification of Emergent, as further explained in Section~\ref{sec:data}. The winner of the associated competition used an ensemble of a convolutional neural network on word embeddings of both the head and body texts, and gradient boosted decision trees with external input features \cite{baird_talos_2017}. The second team proposed an ensemble of five separately trained, randomly initialized MLP's, with solely external input features \cite{hanselowski_description_2017}. A team from UCL finishes third with a model they describe as a `simple but tough-to-beat baseline', using a feature vector consisting only of two 5,000-dimensional term frequency vectors, of the head and body respectively, and the TF-IDF cosine similarity of the two vectors, passed through an MLP with one layer \cite{riedel_simple_2017}. Although these teams achieve impressive overall accuracy (>88\%), it should be noted that their performance is highly unbalanced. They all have accuracy <60\% in the \textit{agree} class and <10\% for \textit{disagree}. 

After the competition finished, numerous researchers have continued using this data set for stance detection, and managed to obtain improved results. \citealp{bhatt_benefit_2017} perform classification combining neural, statistical and external features. Their approach is very similar to that of the UCL team, but they add more external features and include skip-thought sentence embeddings \cite{kiros2015skip}. \citealp{borges_combining_2019} encode head and body using bi-directional RNN's and combine these with numerous similarity features. The current state-of-the-art is set by \citealp{sepulveda-torres_exploring_2021} who apply a large pre-trained language model, specifically RoBERTa \cite{liu2019roberta}. They also use a hierarchical classification approach and in both stages combine RoBERTa embeddings of the head and body with hand-crafted features.

We note that virtually all work on this data set develops classifiers that use as input both hand-crafted (external) features from the head and body texts, and embeddings of these texts, or exclusively the former. Our approach, on the other hand, is based solely on sentence embeddings and does not rely on any hand-crafted representations.
More importantly, although others have introduced methods utilizing large pre-trained language models \cite{sepulveda-torres_exploring_2021, slovikovskaya2019transfer}, they have only applied models that have been trained for generic NLP tasks. To the best of our knowledge, no research is available that attempts to perform stance detection using (and combining) large transformer language models trained on specific tasks relevant to the problem, i.e.\ NLI and SIM. It has been repeatedly shown that such models are tremendously successful in these tasks (see e.g.\ \citealp{liu2019roberta, reimers_sentence-bert_2019, jiang2019evaluating}).

\section{Data set and task} 
\label{sec:data}
As described above, we use the FNC-1 data set for classifying the stance of head-body pairs of news articles on 300 topics as either \textit{agree}, \textit{disagree}, \textit{discuss} or \textit{unrelated} (i.e.\ AGR, DSG, DSC, UNR). The associated stance detection task consists of learning a function that maps a head-body pair to a stance, i.e.\ $f: (h,b) \mapsto s$, with $s\in S = \{AGR, DSG, DSC, UNR\}$. This set-up for stance detection is an extension on the work of \citealp{ferreira_emergent_2016}. Within each topic, headlines and bodies are cross-paired, and for each pairing the appropriate stance is determined by human annotation. Moreover, unrelated head-body samples have been generated by randomly pairing heads with bodies from different topics. Approximately two thirds of the data is contained in the training set and the rest is saved for the test set. More details on how the data set has been constructed can be found on the website of FNC-1. An overview of the set size and distribution of stances is provided in Table~\ref{tab:data set}. 
\begin{table}[h]
    \centering
    \begin{tabular}{cccccc}
         \FL Articles &Samples &AGR &DSG &DSC &UNR  
         \ML 2,587 &75,385 &7.4\% &2.0\% &17.7\% &72.8\%
         \LL
    \end{tabular}
    \caption{Size and stance distribution of FNC-1 complete data set (training and test). Each article consists of one head and body.}
    \label{tab:data set}
\end{table}

Crucially, it should be noted that the data is heavily unbalanced. The \textit{unrelated} samples constitute over 70\% of the data, while only 2\% of samples are labelled as \textit{disagree}. This could cause significant problems in training of the models. In Section~\ref{sec:method data aug} we discuss several methods to alleviate such issues. 

To perform the stance classification task, we firstly pre-process the data by tokenizing the body texts into sentences. This is done using the \verb+sent_tokenize+ function from the \verb+nltk.tokenize+ package of NLTK, a well-known Python library for NLP \cite{bird2004nltk}.

\section{Methodology}
\label{sec:methodology}
We propose a method based on applying large transformer language models specifically pre-trained on NLP tasks that are relevant to FNC-1. We use models from \textit{SentenceTransformers} (aka sBERT), a Python framework providing sentence-level BERT-based models trained on various tasks \cite{reimers_sentence-bert_2019}. Specifically, the NLI embeddings are generated using a \textit{RoBERTa} model \cite{liu2019roberta} pre-trained on a number of NLI tasks\footnote{\url{https://huggingface.co/sentence-transformers/nli-distilroberta-base-v2}}, and the SIM embeddings are generated using a \textit{MiniLM} \cite{wang2020minilm} model trained with a self-supervised contrastive learning objective to identify similar sentences on a large number of sentence-level datasets\footnote{\url{https://huggingface.co/sentence-transformers/all-MiniLM-L6-v2}}. Further specifications of the models and training procedures can be found in the sBERT paper \cite{reimers_sentence-bert_2019} and website\footnote{\url{https://www.sbert.net/index.html}}. In our training procedure, we have elected to completely `freeze' both pre-trained models. This allows us to compute the NLI and SIM sentence embeddings for all heads and bodies as a pre-processing step, which massively reduces training time, and ultimately the size of our models. We denote these four sets of embeddings by SIM-head, NLI-head, SIM-body, and NLI-body, where the former two consist of one embedding each and the latter two consist of a number of embeddings equal to the number of sentences in the body. These four embedding sets constitute the only inputs used in our models. To enable efficient batch processing during training time, we set a fixed body length of 50. Shorter bodies are padded with all-zero embeddings and longer bodies are truncated to include only the first 50 embeddings. The number 50 was chosen based on the distribution of body lengths, a histogram for which is included in the Appendix (Figure~\ref{fig: hist num sents}). 

As described in the introduction, we adopt a hierarchical classification approach. That is, given a head-body pair, we first predict whether they are \textit{unrelated} or not, the model for which will be described in Section~\ref{sec: clas1}. For the second stage of classification, i.e.\ distinguishing between related stances (\textit{agree, disagree, discuss}), we develop two similar models, as described in Section~\ref{sec: clas2}. We then concisely describe how these two stages are combined in Section~\ref{sec: clas3} and finally, in Section~\ref{sec:method data aug}, we propose some data augmentation strategies.

\subsection{Classifying Related-Unrelated: RelatedNet} \label{sec: clas1}

For the first classification stage, i.e.\ determining whether a given head is at all related to a given article body, we propose a simple model based solely on the SIM embeddings of the two texts. Specifically, we compute the cosine similarity between SIM-head and each of the SIM-body embeddings. Then, we select the $k$ most similar embeddings, concatenate these with SIM-head, and pass them through an MLP with four hidden layers with dropout \cite{srivastava2014dropout} and ReLU activation functions, and a final Softmax layer for binary classification. 

\begin{figure}
    \centering
    \includegraphics[scale=0.5]{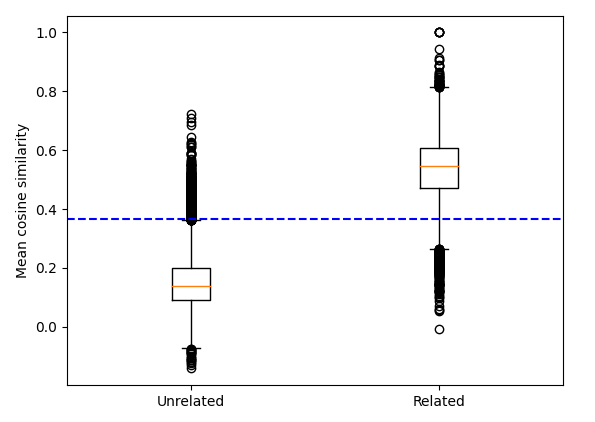}
    \caption{Boxplot of mean cosine similarity between headline and 5 most similar body sentences%
    }
    \label{fig: box sim_scores}
\end{figure}

Even without running any experiments, we have strong reason to believe this method will be successful. Figure~\ref{fig: box sim_scores} shows a boxplot of the mean cosine similarity between SIM-head and  the $k=5$ nearest SIM-body embeddings, for unrelated and related pairs. The blue line shows the classification threshold that maximizes the $F_1$ score, which yields a score of $F_1 = 0.93$. Hence, this result can already be obtained without any training, using only the `cold-start' embeddings of the pre-trained SIM model. The additional MLP classifier should enable further improvement of this result. 

\subsection{Classifying Agree-Disagree-Discuss} \label{sec: clas2}
To address the task of the second classification stage, where we assume the input head-body pair to be related, we propose two models: Top-kNet and AgreemNet. Both are based on the following reasoning. By comparing the SIM-head embedding to the SIM-body embeddings, we can infer which body sentences are most similar to the head, and are thus most informative of the relation between the body and head. Then, we can consider the NLI-body embeddings of these sentences and the NLI-head embedding to classify the stance. The two crucial underlying assumptions of this method are (i) that body sentences which convey the central message of the article are close to the head in SIM embedding space, even though these sentences may contradict each other, and (ii) that when considering these strongly related sentences, their NLI embeddings will contain relevant information to determine their specific relation. To support these assumptions and therefore underpin the argument for our models in general, consider the following head and body sentences, where the first two come from a training sample pair that is marked \textit{agree} and the third from a \textit{disagree} pair. 

H: \textit{Hundreds of Palestinians flee floods in Gaza as Israel opens dams.}\\
B1: \textit{Hundreds of Palestinians were evacuated from their homes Sunday morning after Israeli authorities opened a number of dams near the border, flooding the Gaza Valley in the wake of a recent severe winter storm.}\\
B2: \textit{Locals have continued to use it to dispose of their waste for lack of other ways to do so, however, creating an environmental hazard.}\\
B3: \textit{Easily the most important part of this story is the fact that there are no such dams in Gaza.}

Table~\ref{tab: cos scores} shows the cosine similarity between the embeddings of the head and each of the body sentences in both NLI and SIM space. We observe that SIM embeddings correctly inform us that sentences B1 and B3 are (relatively) important w.r.t.\ the headline, whereas B2 is not. Moreover, in NLI embedding space, the headline is close to B1, but quite far from B3, corresponding to the respective \textit{agree} and \textit{disagree} labels.

\begin{table}[h]
    \centering
    \begin{tabular}{ccc}
         \FL & SIM & NLI
         \ML H, B1 & 0.802 & 0.826
         \NN H, B2 & 0.087 & 0.156
         \NN H, B3 & 0.632 & 0.389
         \LL
    \end{tabular}
    \caption{Cosine similarity between SIM and NLI embeddings of a head and three body sentences}
    \label{tab: cos scores}
\end{table}

\subsubsection{Top-kNet: Most Similar Sentences}
\begin{figure*}
  \centering
  \includegraphics[width=0.9\textwidth]{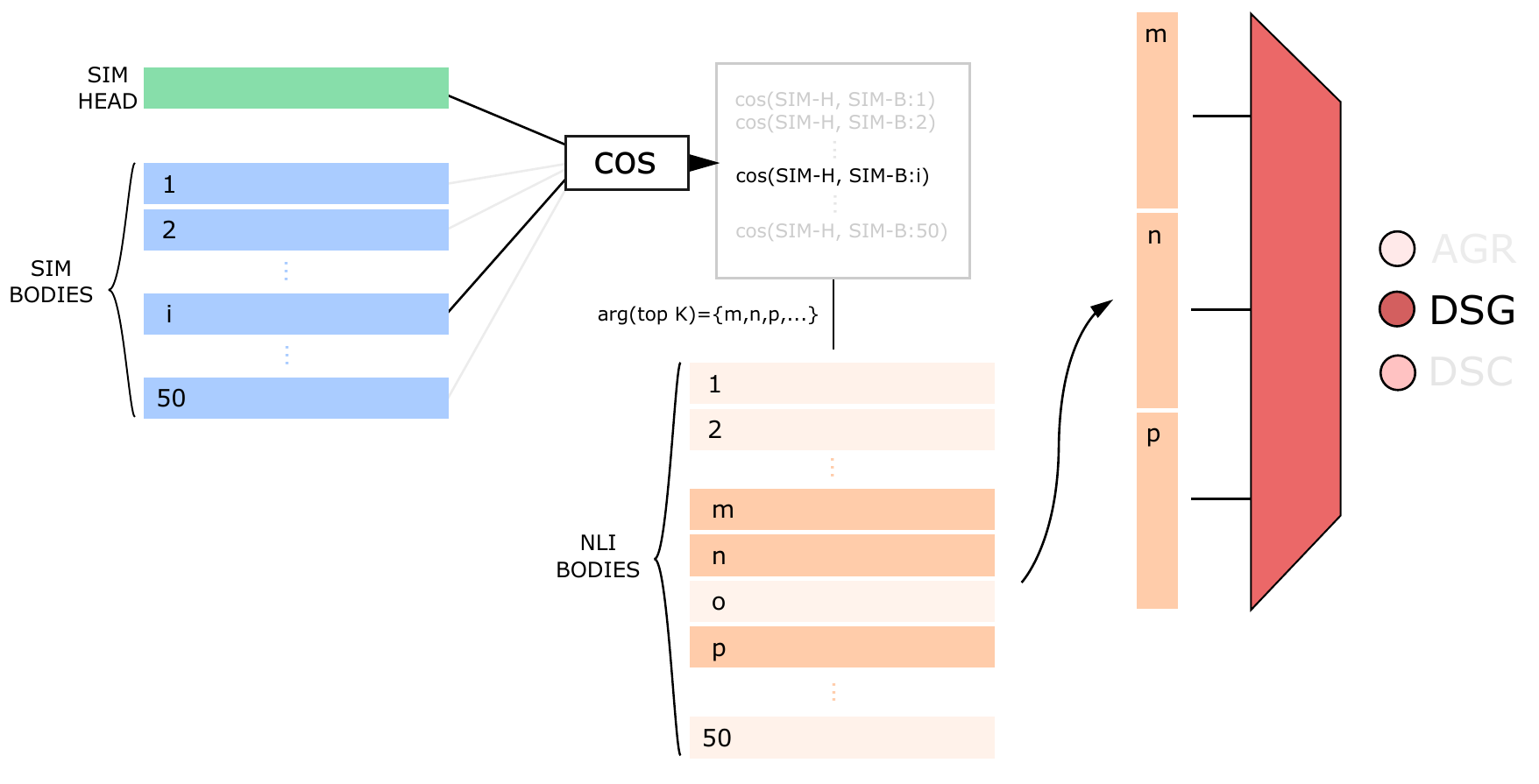}
  \caption{Schematic for TopKNet model}
\label{fig:topk}
\end{figure*}

This model operates by first finding the $k$ body sentences that are most similar to the head, and then using a concatenation of the NLI embeddings for those sentences as input to a Multi-layer Perceptron (MLP) to perform the \textit{Agree-Disagree-Discuss} classification task. The MLP classifier consists of 5 fully-connected layers, with dropout \cite{srivastava2014dropout} and ReLU activations used on each of the hidden layers. The dimensions for the first three hidden layers are controlled by a hyper-parameter $hdim_a$, and, similarly, the dimension of the final hidden layer is controlled by $hdim_b$. The other hyper-parameters for this model are the dropout probability, and $k$, the number of body sentences selected for the classifier. Figure \ref{fig:topk} illustrates the TopKNet architecture.

\subsubsection{AgreemNet: Averaging Over Sentences}
\begin{figure*}
  \centering
  \includegraphics[width=0.7\textwidth]{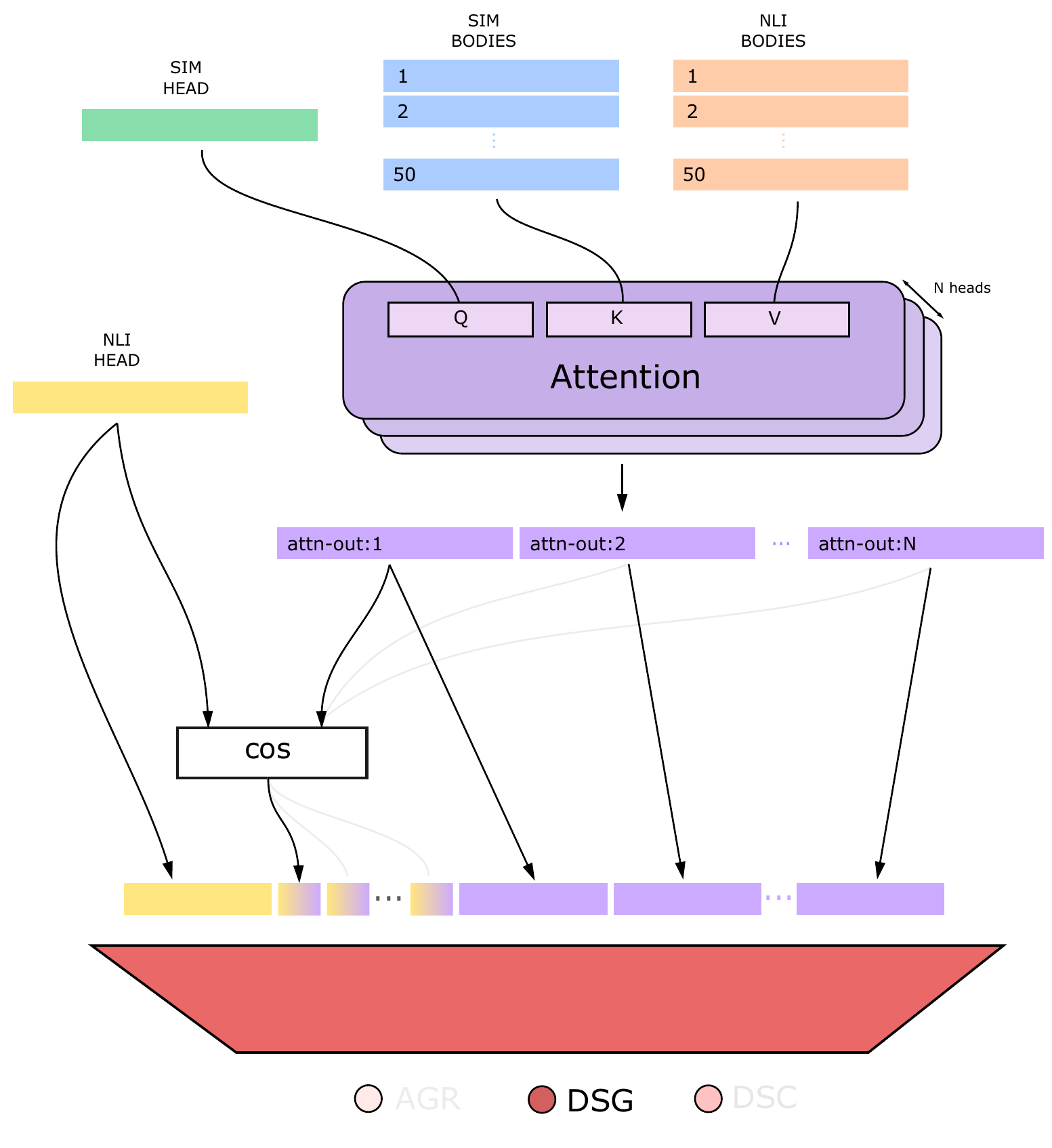}
  \caption{Schematic for AgreemNet model}
\label{fig:agreemnet}
\end{figure*}

This model takes a different approach, computing a similarity-weighted average of NLI-body embeddings as a representation of the body. This is done as in equation \ref{eq:body_avg}, where $\alpha_i$ is the similarity score in $[0,1]$ between the $i^{th}$ body sentence and the head sentence, $\bold b_i$ is the NLI embedding for the $i^{th}$ body sentence, and there are $|B|$ sentences in the body in total.
\begin{equation}
    \label{eq:body_avg}
    \bold b = \sum_{i=1}^{|B|} \alpha_i \bold b_i   
\end{equation}
The body representation, $\bold b$, is then a vector in NLI embedding space, that can be directly compared to the NLI embedding of the head. This weighted average is computed using an attention layer, in which the SIM-head embedding is a \textit{query}, the SIM-body embeddings are \textit{keys}, and the NLI-body embeddings are \textit{values}. Additional modelling flexibility is offered by the attention layer through linear transformations of the query, keys, and values, and the output embedding. Multiple body representations of this form are computed in parallel by AgreemNet using the multi-head scaled dot-product attention proposed in \cite{vaswani2017attention}, which has been implemented in \texttt{pytorch}\footnote{\url{https://pytorch.org/docs/stable/generated/torch.nn.MultiheadAttention.html}}. The concatenation of the body embeddings produced by the attention layer, the NLI-head embedding, and the cosine similarities between the NLI-head and each body embedding, is then passed to an MLP for classification. The MLP used in AgreemNet is similar to that of Top-kNet, but uses 4 fully-connected layers, with ReLUs and dropout after the hidden layers. The hyper-parameters for this model are then $hdim_a$, specifying the dimensions of the first two hidden layers, $hdim_b$ of the final hidden layer, the dropout probability, and the number of attention heads. See Figure~\ref{fig:agreemnet} for a schematic illustration of the AgreemNet model architecture.

\subsection{BaIT: Complete Hierarchical Stance Detection} \label{sec: clas3}
\begin{figure}
  \centering
  \includegraphics[width=0.7\columnwidth]{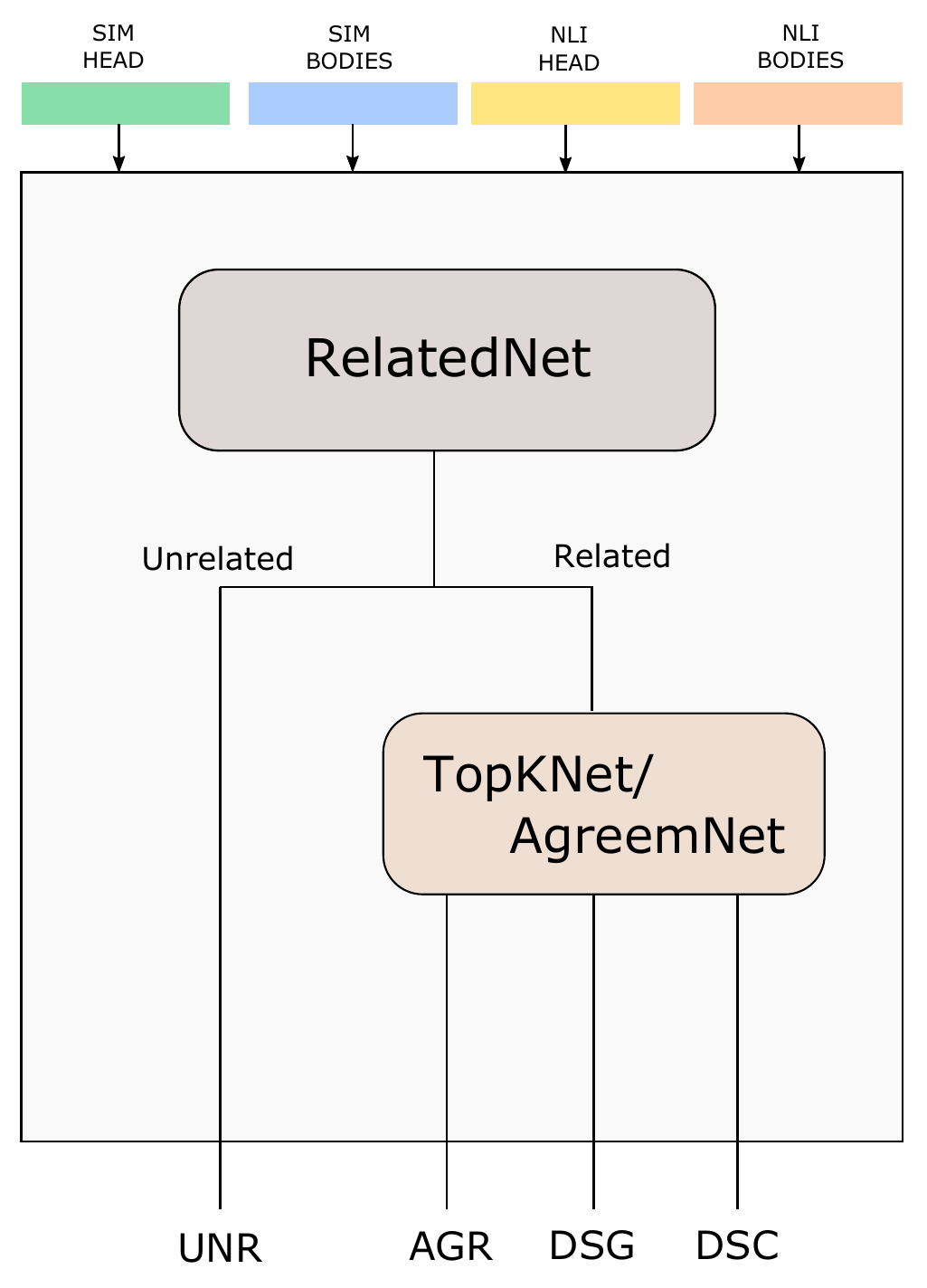}
  \caption{Schematic for hierarchical BaIT model}
\label{fig:bait}
\end{figure}

Finally, to perform complete hierarchical stance detection at test time, we input the SIM and NLI embeddings of a head-body pair into RelatedNet. If this predicts the pair to be \textit{unrelated}, this is returned as output and the procedure is terminated. Otherwise, the embeddings are fed into either AgreemNet or Top-kNet to obtain a final stance prediction of either \textit{agree}, \textit{disagree} or \textit{discuss}. We referred to this combined model as BaIT---the `Barometer for Information Trustworthiness'---and an overview of the architecture is presented in Figure~\ref{fig:bait}. As described above, the models for the two stages are trained separately, and the complete BaIT model is only applied at test time. 

\subsection{Data Augmentation}
\label{sec:method data aug}
In order to combat the imbalanced nature of the FNC-1 data set, a number of data augmentation strategies were employed. The aim in applying these methods is mainly to improve accuracy on the under-represented \textit{disagree} class, thus improving the recall on news articles that spread fake news.

\subsubsection{Weighted Loss Function}
\label{sec:meth:weighted-loss}
This method involves adding a class-dependent weighting to the loss function so that mis-classifying some classes incurs more loss than others. Assigning a higher weighting to under-represented classes in the dataset then helps prevent the model from over-predicting common classes, resulting in more balanced per-class accuracies. The class weights were computed using scikit-learn's \cite{pedregosa2011scikit} \texttt{compute\_class\_weight} function \footnote{\url{https://scikit-learn.org/stable/modules/generated/sklearn.utils.class_weight.compute_class_weight.html}}.

\subsubsection{Synthesis of Negated Headlines}
\label{sec:meth:neg-synth}
An algorithm was developed to semantically negate headlines with the \textit{agree} or \textit{disagree} label, such that the correct label would flip, enabling us to generate more \textit{disagree} samples from existing \textit{agree} samples. This algorithm operates by sequentially attempting three methods of sentence negation, and returning the negated sentence produced by the first applicable method. The Stanford CoreNLP package was used for dependency parsing \cite{chen2014fast}. The first method checks for the presence of the word \textit{not}. If this \textit{not} is functioning as a negation modifier \cite{dep2008stanford}, then it is removed. The second method checks if the root verb of the sentence has an auxiliary verb and, if so, adds the word \textit{not} afterwards. The third method finds the set of antonyms for the root verb using WordNet \cite{miller1995wordnet} and swaps the root with the antonym that results in the headline giving the highest language model probability, according to a distilled GPT-2 model \cite{radford2019language}. This method, when applied to the training set, generated an additional 1068 \textit{disagree} samples, more than doubling the number disagree samples for training.

\subsubsection{Extending the train set with ARC}
In a retrospective analysis of the FNC-1 competition, it is suggested to evaluate the robustness and generalizability of the developed models by applying them to a different but similar data set \cite{hanselowski2018retrospective}. Specifically, the authors suggest using the Argument Reasoning Comprehension (ARC) data set, which is constructed from the debate section of the New York Times, consisting of user posts relating to 188 debate topics \cite{habernal2017argument}. To adapt this set to the stance detection task, a user post is taken as the article body and one of two associated opposing claims is taken as the headline. If the post is labelled as supporting the selecting claim, the sample is marked \textit{agree}, if it supports the opposing claim the sample is \textit{disagree}, and when it is labelled as supporting neither the sample is \textit{discuss}. Similar to the FNC-1 set, \textit{unrelated} samples are generated by pairing user posts with claims from different topics. An overview of this data set is presented in Table~\ref{tab:ARC set}. Although there is again a majority of \textit{unrelated} samples, the other stances have a much more balanced representation, containing substantially more \textit{disagree} samples. 

We investigate if we can improve the performance of our models by additional training on the ARC set. Although this implies additional training samples of the previously under-represented classes, it should be noted that the language used in the ARC set differs significantly from the FNC-1 set. The claims are generally much more factual, simpler sentences rather than news headlines (e.g.\ "\textit{Same-sex colleges are still relevant}" vs "\textit{The Greater Gaza Plan: Is Israel trying to force Palestinians into Sinai?}"), and the users posting to a debate section write differently than journalists. The ARC set is similar enough to FNC, however, that we believe training on a new dataset consisting of both FNC and ARC can help improve performance on the FNC task, particularly on the \textit{disagree} class, which is much better represented in ARC. Note that this augmentation method, unlike the two methods described in Sections \ref{sec:meth:weighted-loss} and \ref{sec:meth:neg-synth}, uses external data, and therefore does not qualify for the FNC-1 challenge.

\begin{table}[h]
    \centering
    \begin{tabular}{cccccc}
         \FL Articles &Samples &AGR &DSG &DSC &UNR  
         \ML 4,488 &17,792 &8.9\% &10.0\% &6.1\% &75.0\%
         \LL
    \end{tabular}
    \caption{Size and stance distribution of ARC complete data set (training and test). Each article consists of one head and body.}
    \label{tab:ARC set}
\end{table}

\section{Experiments}
A number of experiments were carried out to evaluate the performance of the models proposed in Sections~\ref{sec: clas1} and \ref{sec: clas2}. Details on the hyper-parameter tuning process and initial results on the test set are provided in Section~\ref{sec:exp:model-selection}. Further experimental results provided in Section~\ref{sec:exp:data-augmentation} evaluate the effectiveness of each data augmentation method described in Section~\ref{sec:method data aug}.

\label{sec:experiments}

\subsection{Model Selection}
\label{sec:exp:model-selection}
\begin{figure}
  \centering
  \includegraphics[width=0.85\columnwidth]{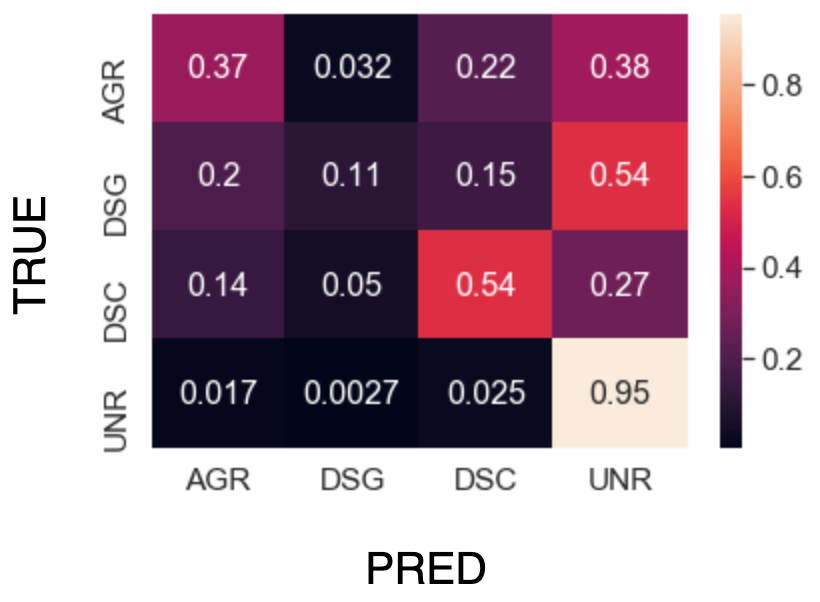}
  \caption{Confusion matrix for BaIT$_K$ model.}
\label{fig:bait_k_cm}
\end{figure}

\begin{figure}
  \centering
  \includegraphics[width=0.85\columnwidth]{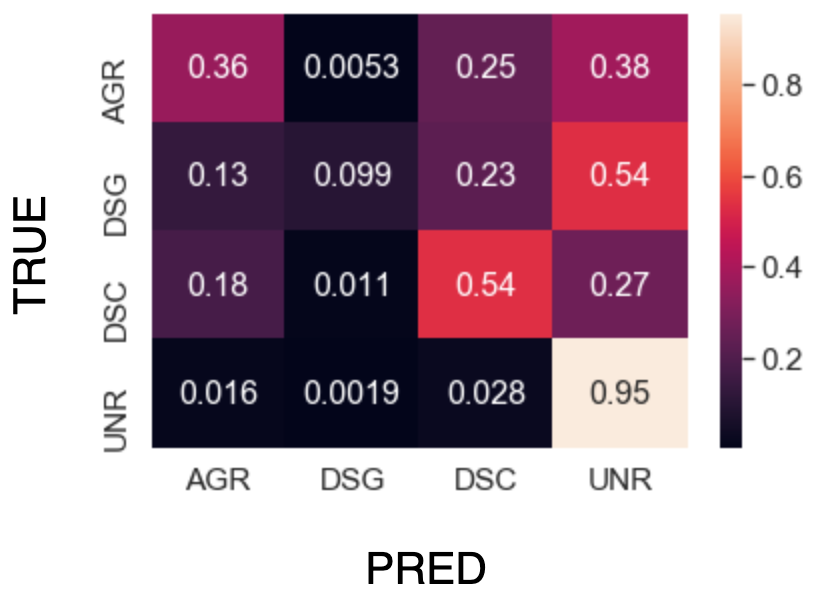}
  \caption{Confusion matrix for BaIT$_A$ model.}
\label{fig:bait_a_cm}
\end{figure}

The FNC-1 training set was split into two distinct subsets for the purpose of hyper-parameter tuning: a train set, and a validation set. The validation set was created by taking samples associated with a selected 30\% of the set of training headlines, such that there is no overlap in headlines between the train and validation set. Hyper-parameter tuning was then carried out using Bayesian optimization, in which the relationship between the hyper-parameters and the unweighted average class accuracy is modelled as a Gaussian Process, with an \textit{acquisition function} choosing new hyper-parameter values that are likely to improve the performance metric \cite{snoek2012bayesian-optimize}. In addition to the hyper-parameters mentioned for each model in Section~\ref{sec:methodology}, the batch size and learning rate were also tuned. The parameter values selected for each model by the Bayesian optimization procedure are presented in Table \ref{tab:hyperparam-results}. Using these hyper-parameters, the BaIT model was trained on the full FNC-1 training set and then evaluated on the test set using both TopKNet and AgreemNet (BaIT$_K$ and BaIT$_A$, respectively) for comparison. Evaluation results are reported in Table \ref{tab:fnc-results}, where they are presented alongside the results from the Simple but Tough To Beat (STTB) model \cite{riedel_simple_2017}, which came 3rd on the official FNC-1 leaderboard. Confusion matrices for both BaIT$_K$ and BaIT$_A$ are presented in Figures \ref{fig:bait_k_cm} and \ref{fig:bait_a_cm}. It is noteworthy that while the overall accuracy of STTB is higher, BaIT$_K$ and BaIT$_A$ achieve a higher accuracy on the \textit{disagree} class, which is integral to identifying fake news in this task. Note also that BaIT$_K$ has $\sim$5 times fewer parameters than STTB: BaIT$_K$ has 195,543 parameters, while STTB has 1,000,500.

\begin{table}[h]
    \centering
    \begin{tabular}{c|ccc}
         \FL                    & RelatedNet    & TopKNet   & AgreemNet
         \ML Batch size         & 32            & 64        & 128
         \NN Learning rate      & $10^{-4}$        & $10^{-3}$     & $10^{-3}$ 
         \NN Dropout            & 0.277         & 0.301     & 0.105
         \ML $hdim_a$           & 600           & 60        & 60
         \NN $hdim_b$           & 600           & 60        & 20
         \NN $k$                & 4             & 3         & $-$
         \NN Attn heads    & $-$           & $-$       & 11
         \ML Params. & $2,235,602$ & $195,543$ & $2,203,679$
         \LL
    \end{tabular}
    \caption{Optimized hyper-parameters for each model.}
    \label{tab:hyperparam-results}
\end{table}

\begin{table}[h]
    \centering
    \begin{tabular}{c|ccc}
        \FL             & BaIT$_K$  & BaIT$_A$  & STTB 
         \ML  AGR        & 36.8     & 36.5     & 44.0
         \NN  DSG       & 11.3     & 9.9       & 6.6
         \NN  DSC       & 53.6     & 53.9     & 81.4
         \NN  UNR       & 95.5     & 95.5     & 97.9
         \ML  Overall   & 81.4     &  81.4    & 88.5
         \NN  FNC-1     & 68.2    &  68.2    & 81.7 
         \LL
    \end{tabular}
    \caption{Per-class accuracy (\%), overall accuracy (\%), and FNC-1 score for our models and the STTB model \cite{riedel_simple_2017}.}
    \label{tab:fnc-results}
\end{table}

\subsection{Data Augmentation}
\label{sec:exp:data-augmentation}
Each data augmentation method described in Section~\ref{sec:method data aug} was applied to both BaIT$_K$ and BaIT$_A$ models, using the optimized hyper-parameters specified in Table~\ref{tab:hyperparam-results}. The FNC-1 test scores achieved using these methods are given in Table~\ref{tab:data-aug-results} for the two BaIT variants.
\begin{table*}[!htbp]
    \centering
    \begin{tabular}{c|ccc|ccc}
        \FL & \multicolumn{3}{c|}{BaIT$_A$} & \multicolumn{3}{c}{BaIT$_K$}
        \NN             & Weighted Loss     & Synthetic Data    & +ARC & Weighted Loss     & Synthetic Data    & +ARC 
         \ML  AGR       & 43.5             & 36.5             & 54.7 & 44.1             & 47.4             & 39.5
         \NN  DSG       & 9.8              & 1.0                 & 17.4 & 17.8             & 18.2             & 24.4
         \NN  DSC       & 46.0             & 32.2             & 63.7 & 41.0             & 30.4             & 73.0
         \NN  UNR       & 95.5             & 94.4             & 90.8 & 95.5             & 94.4             & 90.8
         \ML  Overall   & 80.5             & 78.0             & 81.3 & 79.9             & 77.5             & 82.0
         \NN  FNC-1     & 66.8             & 64.4             & 75.6 & 65.7             & 63.6             & 76.7
         \LL
    \end{tabular}
    \caption{Per-class accuracy (\%), overall accuracy (\%), and FNC-1 score for the two BaIT variants, using a number of data augmentation strategies.}
    \label{tab:data-aug-results}
\end{table*}
In general, the data augmentation methods were successful in improving the performance on the under-represented \textit{disagree} class. The most effective augmentation method was including the ARC set in the training data, doubling the \textit{disagree} accuracy for BaIT$_K$ and almost tripling it for BaIT$_A$. Note, however, that this method involves adding external data, and so is in general far more costly than the other methods. The next biggest improvement resulted from using synthetic data with BaIT$_A$, which caused a doubling in performance. Weighting the \textit{disgree} loss also hugely improved performance here, again almost doubling the \textit{disagree} accuracy. Note that while the weighted loss resulted in slightly lower \textit{disagree} accuracy than synthetic data, it had less negative impact on the overall accuracy, reducing it by $\sim$1\%, rather than $\sim$3\%. Interestingly, the weighted loss and synthetic data methods caused a decrease in the \textit{disagree} accuracy in the BaIT$_K$ model, perhaps due to a lack of modelling flexibility resulting from its small size. Using these methods, we can achieve 3 times the \textit{disagree} accuracy of STTB with only 7\% less overall accuracy, by using synthetic data and the BaIT$_K$ model, which only requires the original FNC-1 training data and thus fits the original criteria for the challenge. By extending the dataset using ARC, this improvement grows to 4 times the \textit{disagree} accuracy of STTB.

\section{Conclusion}
\label{sec:conclusion}
This paper proposes new method of approaching the FNC-1 task using pre-trained encoders, which greatly reduces the number of trainable parameters required. Two new architectures using this approach are proposed, one simple smaller model, BaIT$_K$, and one larger more complex model, BaIT$_A$, both of which achieve a performance competitive with existing baselines, particularly on the \textit{disagree} class, which is key to this challenge. A number of methods in data augmentation are also explored, in attempt to further improve \textit{disagree} performance and overcome class imbalance in the dataset. This included the proposal of a novel data augmentation method, involving an algorithm that generates negated versions of headlines to produce new samples of the under-represented class. These methods were broadly successful, particularly when used with the more flexible BaIT$_A$ model. Future work could look towards experimenting with other SIM and NLI pre-trained models to compare effectiveness of knowledge transfer to the FNC-1 task. Additionally, more sophisticated methods in negated sentence generation could further tackle class imbalance in such datasets.

\bibliography{refs}

\section{Appendix}
All code can be found at \url{https://github.com/OisinNolan/fakenewschallenge}.

\begin{figure}[H]
    \centering
    \includegraphics[scale =0.5]{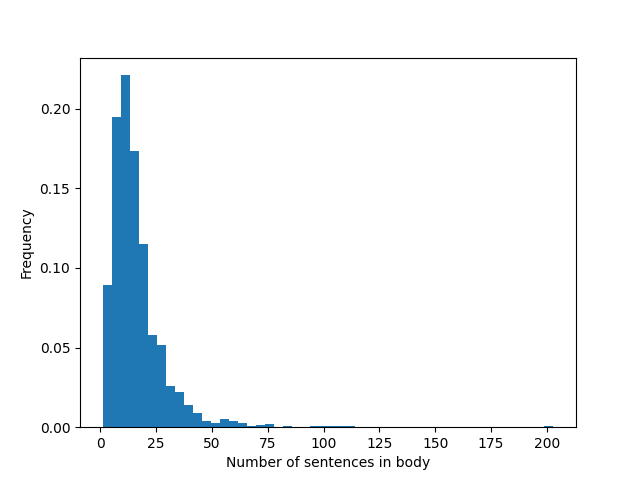}
    \caption{Histogram of number of sentences in article bodies}
    \label{fig: hist num sents}
\end{figure}

\end{document}